\documentclass[twoside]{article}
\usepackage[accepted]{aistats2016}

\usepackage{latexsym, balance}
\usepackage{amsmath}
\usepackage{multirow}
\usepackage{url}
\usepackage{graphicx}
\usepackage{subfigure}
\usepackage{algorithm}
\usepackage{algorithmic,xpatch}
\usepackage{multirow}
\usepackage{color}
\usepackage{url}

\newtheorem{lemma}{Lemma}

\usepackage{graphicx}

\xpatchcmd{\algorithmic}{\setcounter}{\algorithmicfont\setcounter}{}{}
\providecommand{\algorithmicfont}{}
\providecommand{\setalgorithmicfont}[1]{\renewcommand{\algorithmicfont}{#1}}

\def \XPANDR{\textsc{EXPANDER}}

\def \DXPANDR{{\textsc{DIST-EXPANDER}}}

\def \FREQ{\textsc{FREQ-THRESH}}

\begin{document}

\runningauthor{Sujith Ravi, Qiming Diao}

\twocolumn[

\aistatstitle{Large Scale Distributed Semi-Supervised Learning Using Streaming Approximation}

\aistatsauthor{ Sujith Ravi \And Qiming Diao\footnotemark} 
\aistatsaddress{ Google Inc., Mountain View, CA, USA \\{\tt sravi@google.com} \And Carnegie Mellon University, Pittsburgh, PA, USA \\ Singapore Mgt. University, Singapore \\ {\tt qiming.ustc@gmail.com}} ]

\footnotetext{Work done during an internship at Google.}

\begin{abstract}
Traditional graph-based semi-supervised learning (SSL) approaches are not suited for massive data and large label scenarios since they scale linearly with the number of edges $|E|$ and distinct labels $m$. To deal with the large label size problem, recent works propose sketch-based methods to approximate the label distribution per node thereby achieving a space reduction from $O(m)$ to $O(\log m)$, under certain conditions. In this paper, we present a novel streaming graph-based SSL approximation that effectively captures the sparsity of the label distribution and further reduces the space complexity per node to $O(1)$. We also provide a distributed version of the algorithm that scales well to large data sizes. Experiments on real-world datasets demonstrate that the new method achieves better performance than existing state-of-the-art algorithms with significant reduction in memory footprint. Finally, we propose a robust graph augmentation strategy using unsupervised deep learning architectures that yields further significant quality gains for SSL in natural language applications.
\end{abstract}

\setalgorithmicfont{\footnotesize}

\vspace{-.25in}
\section{Introduction}
\label{sec:intro}

Semi-supervised learning (SSL) methods use small amounts of labeled data along with large amounts of unlabeled data to train prediction systems. Such approaches have gained widespread usage in recent years and have been rapidly supplanting supervised systems in many scenarios owing to the abundant amounts of unlabeled data available on the Web and other domains. Annotating and creating labeled training data for many predictions tasks is quite challenging because it is often an expensive and labor-intensive process. On the other hand, unlabeled data is readily available and can be leveraged by SSL approaches to improve the performance of supervised prediction systems.
\vspace{-0.3em}

There are several surveys that cover various SSL methods in the literature~\cite{Seeger01,Zhu05,ChaSchZie06,blitzer-zhu:2008:ACLTutorials}. The majority of SSL algorithms are
computationally expensive; for example, transductive SVM~\cite{Joachims:1999}. Graph-based SSL
algorithms~\cite{Zhu03ssl,Joachims03,Tommi03,Belkin05,SubramanyaB09,Talukdar:ACL'10} are a subclass of SSL techniques that have received a lot of attention recently, as they
scale much better to large problems and data sizes. These methods exploit the idea of constructing and smoothing a graph in which data (both labeled and unlabeled) is represented
by nodes and edges link vertices that are related to each other. Edge weights are defined using a similarity function on node pairs and govern how strongly the labels of the
nodes connected by the edge should agree. Graph-based methods based on label propagation~\cite{Zhu03ssl,Talukdar:ECML'09} work by using class label information associated with
each labeled ``seed'' node, and propagating these labels over the graph in a principled, iterative manner. These methods often converge quickly and their time and space
complexity scales linearly with the number of edges $|E|$ and number of labels $m$. Successful applications include a wide range of tasks in computer vision~\cite{WangJC13}, information retrieval (IR) and social networks~\cite{Ugander:2013} and natural language processing (NLP); for example, class instance acquisition and relation prediction, to name a few~\cite{Talukdar:ACL'10,Subramanya:EMNLP'10,KozarevaVT11}.
\vspace{-0.3em}

Several classification and knowledge expansion type of problems involve a large number of labels in real-world scenarios. For instance, entity-relation classification over the widely used Freebase taxonomy requires learning over thousands of labels which can grow further by orders when extending to open-domain extraction from the Web or social media; scenarios involving complex overlapping classes~\cite{CarlsonBKSHM10}; or fine-grained classification at large scale for natural language and computer vision applications~\cite{TalukdarC14,imagenet_cvpr09}. Unfortunately, existing graph-based SSL methods cannot deal with large $m$ and $|E|$ sizes. Typically individual nodes are initialized with sparse label distributions, but they
become dense in later iterations as they propagate through the graph. Talukdar and Cohen~\shortcite{TalukdarC14} recently proposed a method that seeks to overcome the label scale
problem by using a Count-Min Sketch~\cite{Cormode:2005} to approximate labels and their scores for each node. This reduces the memory complexity to $O(\log m)$ from $O(m)$. They
also report improved running times when using the sketch-based approach. However, in real-world
applications, the number of {\it actual} labels $k$ associated with each node is typically sparse even though the overall label space may be huge; i.e., $k \ll m$. Cleverly
leveraging sparsity in such scenarios can yield huge benefits in terms of efficiency and scalability. While the sketching technique from~\cite{TalukdarC14} approximates the label
space succinctly, it does not utilize the sparsity (a naturally occurring phenomenon in real data) to full benefit during learning.

\noindent {\bf Contributions:} In this paper, we propose a new graph propagation algorithm for general purpose semi-supervised learning with applications for NLP and other areas. We show how the new algorithm can be run efficiently even when the label size $m$ is huge. At its core, we use an approximation that effectively captures the sparsity of the label distribution
and ensures the algorithm propagates the labels accurately. This reduces the space complexity per node from $O(m)$ to $O(k)$, where $k$$\ll$$m$ and a constant (say, 5 or 10 in practice), so $O(1)$ which scales better than previous methods. We show how to efficiently parallelize the algorithm by proposing a distributed version that scales well for large graph sizes. We also propose an efficient linear-time graph construction strategy that can effectively combine information from multiple signals which can vary between {\it sparse} or {\it dense} representations. In particular, we show that for graphs where nodes represent textual information (e.g., entity name or type), it is possible to robustly learn latent semantic embeddings associated with these nodes using only raw text and state-of-the-art deep learning techniques. Augmenting the original graph with such embeddings followed by graph SSL yields significant improvements in quality. We demonstrate the power of the new method by evaluating on different knowledge expansion tasks using existing benchmark datasets. Our results show that, when compared with existing state-of-the-art systems for these tasks, our method performs better in terms of space complexity and qualitative performance.

\section{Graph-based Semi-Supervised Learning}
\label{sec:method}

\noindent {\bf Preliminary:}
The goal is to produce a soft assignment of labels to each node in a graph
$G=(V,E,W)$, where $V$ is the set of nodes, $E$ the set of edges and $W$ the edge weight matrix.\footnote{The graph $G$ can be directed or undirected depending on the task. Following most existing works in the literature, we use undirected edges for $E$ in our experiments.} Every edge $(v,u) \notin E$ is assigned a weight $w_{vu}=0$. Among the $|V|=n$ number of nodes, $|V_l|=n_l$ of
them are labeled while $|V_u|=n_u$ are unlabeled. We use diagonal matrix $S$
to record the seeds, in which $s_{vv}=1$ if the node $v$ is seed. $L$ represents the output label set whose size $|L| = m$ can be large in the real world. $\bf Y$ is a $n*m$ matrix which records the training label distribution for the seeds where $Y_{vl}=0$ for $v \in V_u$, and $\bf \hat{Y}$ is an $n*m$ label distribution assignment matrix for all nodes. In general, our method is a graph-based semi-supervised learning algorithm, which learns $\bf \hat{Y}$ by propagating the information of $\bf Y$ on graph $G$.

\subsection{Graph SSL Optimization}

We learn a label distribution $\hat{Y}$ by minimizing the convex objective function:
\begin{small}
\begin{align}
\mathcal{C}(\bf \hat{Y}) &= \mu_1 \sum_{v \in V_l}s_{vv}||{\bf \hat{Y}_v} - {\bf Y_v}||_2^2 \notag\\
     &+ \mu_2 \sum_{v\in V, u\in \mathcal{N}(v)}w_{vu} ||{\bf \hat{Y}_v} - {\bf \hat{Y}_u}||^2 \notag \\
     &+ \mu_3 \sum_{v \in V}||{\bf \hat{Y}_v} - {\bf U}||_2^2 \notag \\
     &s.t. \sum_{l=1}^L \hat{Y}_{vl} = 1, \forall v
\label{align:obj}
\end{align}
\end{small}
where $\mathcal{N}(v)$ is the (incoming) neighbor node set of the node $v$, and $U$ is
the (uniform) prior distribution over all labels. The above objective function models that: 1) the label distribution should be close to the gold label
assignment for all the seeds; 2) the label distribution of a pair of neighbors
should be similar measured by their affinity score in the edge weight matrix; 3) the label
distribution should be close to the prior $\bf U$, which is a uniform
distribution. The setting of the hyperparameters $\mu_i$ will be discussed in
Section~\ref{sec:expsetup}.

The optimization criterion is inspired from~\cite{Bengio+al-ssl-2006} and similar to some existing approaches such as Adsorption~\cite{Baluja:WWW'08} and MAD~\cite{Talukdar:ECML'09} but uses a slightly different objective function, notably the matrices have different constructions. In Section~\ref{sec:exp}, we also compare our vanilla version against some of these baselines for completeness.

\label{sec:algo}

The objective function in Equation~\ref{align:obj} permits an efficient iterative optimization technique that is repeated until convergence. We utilize the Jacobi iterative algorithm which defines the approximate solution at the $(i+1)th$ iteration, given the solution of the $(i)th$ iteration as follows:
\vspace{-0.3in}

\begin{small}
\begin{align}
\begin{split}
\hat{Y}_{vl}^{(i)} &= \frac{1}{M_{vl}}(\mu_1 s_{vv}Y_{vl} + \mu_2 \sum_{u \in \mathcal{N}(v)} w_{vu} \hat{Y}_{ul}^{(i-1)} + \mu_3 U_l)\\
M_{vl} &= \mu_1s_{vv} + \mu_2 \sum_{u \in \mathcal{N}(v)} w_{vu} + \mu_3
\end{split}
\label{align:iterate}
\end{align}
\end{small}
\vspace{-0.2in}

where $i$ is the iteration index and $U_l=\frac{1}{m}$ which is the uniform
distribution on label $l$. The iterative procedure starts with
$\hat{Y}_{vl}^{(0)}$ which is initialized with seed label weight $Y_{vl}$ if
$v \in V_l$, else with uniform distribution $\frac{1}{m}$. In each iteration
$i$, $\hat{Y}_{vl}^{(i)}$ aggregates the label distribution $\bf
\hat{Y}_{u}^{(i-1)}$ at iteration $i-1$ from all its neighbors $u \in
\mathcal{N}(v)$. More details for deriving the update equation can be found in~\cite{Bengio+al-ssl-2006}.

We use the name ~\XPANDR~to refer to this vanilla method that optimizes Equation~\ref{align:obj}.

\subsection{{\small \DXPANDR:} Scaling To Large Data}
\label{sec:distributed}

In many applications, semi-supervised learning becomes challenging when the graphs become huge. To scale to really large data sizes, we propose \DXPANDR, a distributed version of the algorithm that is directly suited towards
parallelization across many machines. We turn to Pregel~\cite{Pregel} and its open source version Giraph~\cite{Giraph} as the underlying framework for our distributed algorithm. These systems follow a Bulk Synchronous Parallel (BSP) model of computation that  proceeds in rounds. In every round, every machine does some local processing and then sends arbitrary messages to other machines. Semantically, we think of the communication graph as fixed, and in each round each node performs some local computation and then sends messages to its neighbors. 

The specific systems like Pregel and Giraph build infrastructure that ensures that the overall system is fault tolerant, efficient, and fast. The programmer's job is simply to specify the code that each vertex will run at every round. Previously, some works have explored using MapReduce framework to scale to large graphs~\cite{TalukdarRPRBP08}. But unlike these methods, the Pregel-based model is far more efficient and better suited for graph algorithms that fit the iterative optimization scheme for SSL algorithms. Pregel keeps vertices and edges on the machine that performs computation, and uses network transfers only for messages. MapReduce, however, is essentially functional, so expressing a graph algorithm as a chained MapReduce requires passing the entire state of the graph from one stage to the next---in general requiring much more communication and associated serialization overhead which results in significant network cost (refer~\cite{Pregel} for a detailed comparison). In addition, the need to coordinate the steps of a chained MapReduce adds programming complexity that is avoided by \DXPANDR~iterations over rounds/steps. Furthermore, we use a version of Pregel that allows spilling to disk instead of storing the entire computation state in RAM unlike~\cite{Pregel}.
Algorithm~\ref{algo:distributed} describes the details.

\setlength{\textfloatsep}{2pt}
\begin{small}
\begin{algorithm}[t]
\caption{$\DXPANDR$ Algorithm}
\begin{algorithmic}[1]
\STATE {\bf Input:} A graph $G = (V, E, W)$, where $V = V_l \cup
V_u$\\\hspace{4pt}$V_l$ = seed/labeled nodes, $V_u$ = unlabeled
nodes \STATE {\bf Output:} A label distribution ${\bf \hat{Y}_v} =
\hat{Y}_{v1} \hat{Y}_{v2} ... \hat{Y}_{vm}$ for every node $v \in V$
minimizing the overall objective function~(\ref{align:obj}). Here,
$\hat{Y}_{vl}$ represents the weight of label $l$ assigned to the node $v$.
\STATE Let $L$ be the set of all possible labels, $|L| = m$. \STATE Initialize
$\hat{Y}_{vl}^0$ with seed label weights if $v \in V_l$, else $\frac{1}{m}$.
\STATE (Graph Creation) Initialize each node $v$ with its neighbors
$\mathcal{N}(v) = \{ u : (v, u) \in E\}$. \STATE Partition the graph into $p$
disjoint partitions $V_1, ... , V_p$, where $\bigcup_i V_i = V$. \FOR{$i$ = 1 to
$max\_iter$} \STATE Process individual partitions $V_p$ in parallel.
\FOR{every node $v \in V_p$} \STATE (Message Passing) Send previous label distribution ${\bf \hat{Y}_v^{i-1}}$ to all neighbors $u
\in \mathcal{N}(v)$. \STATE (Label Update) Receive a
message $\mathcal{M}_u$ from its neighbor $u$ with corresponding label weights
${\bf \hat{Y}_u^{i-1}}$. Process each message $\mathcal{M}_1 ...
\mathcal{M}_{|\mathcal{N}(v)|}$  and update current label distribution
$\hat{Y}_v^i$ iteratively using Equation~(\ref{align:iterate}). \ENDFOR \ENDFOR
\end{algorithmic}
\label{algo:distributed}
\end{algorithm}
\end{small}

\vspace{-.1em}
\section{Streaming Algorithm for Scaling To Large Label Spaces}
\label{sec:sparsity}
Graph-based SSL methods usually scale linearly with the label size $m$, and require $\mathcal{O}(m)$ space for each node. Talukdar and Cohen~\shortcite{TalukdarC14} proposed to deal with the issue of large label spaces by employing a Count-Min Sketch approximation to store the label distribution of each node. However, we argue that it is not necessary to approximate the whole label distribution for each node, especially for large label sets, because the label distribution of each node is typically sparse and only the top
ranking ones are useful. Moreover, the Count-Min Sketch can even be harmful for the top ranking labels because of its approximation. The authors also mention other related works that attempt to induce sparsity using regularization techniques~\cite{tibshirani96,kowalski09} but for a very different purpose~\cite{DasSmith:2012}. In contrast, our work tackles the exact same problem as~\cite{TalukdarC14} to scale graph-based SSL for large label settings. The method presented here does not attempt to enforce sparsity and instead focuses on efficiently storing and updating label distributions during semi-supervised learning with a streaming approximation. In addition, we also compare (in Section~\ref{sec:exp}) against other relevant graph-based SSL baselines~\cite{Talukdar:ACL'10,ragrawal2013} that use heuristics to discard poorly scored labels and retain only top ranking labels per node out of a large label set.

\noindent {\bf \XPANDR-S Method:} We propose a streaming sparsity approximation algorithm for semi-supervised learning that achieves constant space complexity and huge memory savings over the current state-of-the-art approach (MAD-SKETCH) in addition to significant runtime improvements over the exact version. The method processes messages from neighbors efficiently in a streaming fashion and records a sparse set of top ranking labels for each node and approximate estimate for the remaining. In general, the idea is similar to finding frequent items from data streams, where the item is the label and the streams are messages from neighbors in our case. Our Pregel-based approach (Algorithm~\ref{algo:distributed}) provides a natural framework to implement this idea of processing message streams. We replace the update Step 11 in the algorithm with the new version thereby allowing us to scale to both large label spaces and data using the same framework.

\noindent {\bf Preliminary:} Manku and Motwani~\shortcite{Manku:VLDB'02} presented an algorithm for computing frequency counts
exceeding a user-specified threshold over data streams, and others have applied this algorithm to handle large amounts of data in NLP problems~\cite{Goyal:NAACL'09,VandurmeLall09,VanDurme:ACL'11,Osborne:ACL'14}. The general
idea is that a data stream containing $N$ elements is split into multiple epochs with $\frac{1}{\epsilon}$ elements in each epoch. Thus there are $\epsilon N$ epochs in total,
and each such epoch has an ID starting from $1$. The algorithm processes elements in each epoch sequentially and maintains a list of tuples of the form $(e,f,\Delta)$, where $e$
is an item, $f$ is its reported frequency, and $\Delta$ is the maximum error of the frequency estimation. In current epoch $t$, when an item $e$ comes in, it increments the
frequency count $f$, if the item $e$ is contained in the list of tuples. Otherwise, it creates a new tuple $(e,1,t-1)$. Then, after each epoch, the algorithm filters out the
items whose maximum frequency is small. Specifically, if the epoch $t$ ended, the algorithm deletes all tuples that satisfies the condition $f+\Delta \leq t$. This ensures that
rare items are not retained at the end.

\noindent {\bf Neighbor label distributions as weighted streams:} Intuitively, in our setting, each item is a label and each neighbor is an epoch. For a given node $v$, the
neighbors pass label probability streams to node $v$, where each neighbor $u
\in \mathcal{N}(v)$ is an epoch and the size of epochs is $|\mathcal{N}(v)|$.
We maintain a list of tuples of the form $(l,f,\Delta)$, in which the $l$ is
the label index, $f$ is the weighted probability value, and $\Delta$ is the
maximum error of the weighted probability estimation. For the current neighbor
$u_t$ (say, it is the $t$-th neighbor of $v$, $t \leq |\mathcal{N}(v)|$), the
node $v$ receives the label distribution $\hat{Y}_{u_tl}$ with edge weight
$w_{vu_t}$. The algorithm then does two things: if the label $l$ is currently in
the tuple list, it increments the probability value $f$ by adding
$w_{vu_t}\hat{Y}_{u_tl}$. If not, it creates new tuple of the form
$(l,w_{vu_t}\hat{Y}_{u_tl}, \delta\sum_{i=1}^{t-1}w_{vu_i})$. Here, we use
$\delta$ as a probability threshold (e.g., can be set as uniform distribution
$\frac{1}{m}$), because the value in an item frequency stream is naturally $1$ while ours
is a probability weight. Moreover, each epoch $t$, which is neighbor $u_t$ in our task,
is weighted by the edge weight $w_{vu_t}$ unlike previous settings~\cite{Manku:VLDB'02}. Then, after we receive the message
from the $t$-th neighbor, we filter out the labels whose maximum probability is
small. We delete label $l$, if $f + \Delta \leq
\delta\sum_{i=1}^{t}w_{vu_i}$.

\noindent {\bf Memory-bounded update:} With the given streaming sparsity algorithm, we can ensure that no low
weighted-probability labels are retained after receiving messages from all
neighbors. However, in many cases, we want the number of retained labels to be
bounded by $k$, i.e retain the top-$k$ label based on the probability. In this
case, for a node $v$, each of its neighbors $u \in \mathcal{N}(v)$ just
contains its top-$k$ labels, i.e. $\bf \hat{Y}_u$$= \hat{Y}_{ul_1},
\hat{Y}_{ul_2}, \cdots, \hat{Y}_{ul_k}$. Moreover, we use $\delta_u =
\frac{1.0 - \sum_{i=1}^{k} \hat{Y}_{ul_i}}{m-k}$ to record the average
probability mass of the remaining labels. We then apply the previous streaming
sparsity algorithm. The only difference is that when a label $l$ does not
exists in the current tuple list, it creates a new tuple of the form
$(l,w_{vu_t}\hat{Y}_{u_tl}, \sum_{i=1}^{t-1}w_{vu_i}\delta_{u_i})$.
Intuitively, instead of setting a fixed global $\delta$ as threshold, we vary
the threshold $\delta_{u_i}$ based on the sparsity of the previous seen
neighbors. In each epoch, after receiving messages $\bf \hat{Y}_{u_t}$ from
the current ($t$-th) neighbor, we scan the current tuple list. For each tuple
$(l,f,\Delta)$, we increments its probability value $f$ by adding
$\delta_{u_t}$, if label $l$ is not within the top-$k$ label list of the
current $t$-th neighbor. Then, we filter out label $l$, if $f + \Delta \leq
\sum_{i=1}^{t}w_{vu_i}\delta_{u_i}$. Finally, after receiving messages from
all neighbors, we rank all remaining tuples based on the value $f+\Delta$
within each tuple $(l,f,\Delta)$. This value represents the maximum
weighted-probability estimation. Then we just pick the top-$k$ labels and
record only their probabilities for the current node $v$.
\begin{lemma}
\label{SparsityApprox}
For any node $u \in V$, let $y$ be the un-normalized true label weights and $\hat{y}$ be the estimate given by the streaming sparsity approximation version of \XPANDR~algorithm at any given iteration. Let $N$ be the total number of label entries received from all neighbors of $u$ before aggregation, $d=|\mathcal{N}(u)|$ be the degree of node $u$ and $k$ be the constant number of (non-zero) entries retained in $\hat{y}$ where $N \le k \cdot d$, then (1) the approximation error of the proposed sparsity approximation is bounded in each iteration by $\hat{y}_l \le y_l \le \hat{y}_l + \delta \cdot \frac{N}{k}$ for all labels $l$, (2) the space used by the algorithm at each node is $O(k)$ = O(1).
\end{lemma}

The proof for the first part of the statement can be derived following a similar analysis as~\cite{Manku:VLDB'02} using label weights instead of frequency. At the end of each iteration, the algorithm ensures that labels with low weights are not retained and for the remaining ones, its estimate is close to the exact label weight within an additive factor. The second part of the statement follows  direclty from the fact that each node retains atmost $k$ labels in every iteration. The detailed proof is not included here.

Next, we study various graph construction choices and demonstrate how augmenting the input graph using external information can be beneficial for learning.
\vspace{-.5em}

\section{Graph Construction}
\label{sec:graph_model}

\begin{figure}[ht!]
\centering
\scalebox{0.6}{
\begin{tabular}{cccc}
\includegraphics[width=35mm]{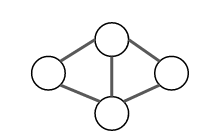} &
      \includegraphics[width=25mm,height=27mm]{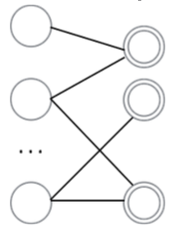} & \includegraphics[width=30mm]{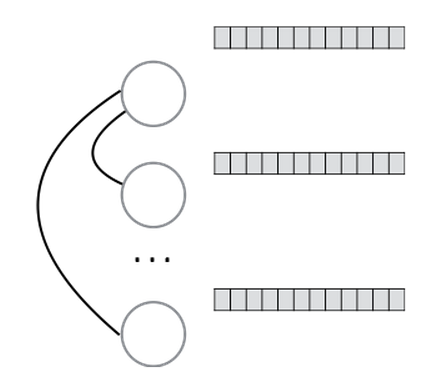} &  \includegraphics[width=30mm]{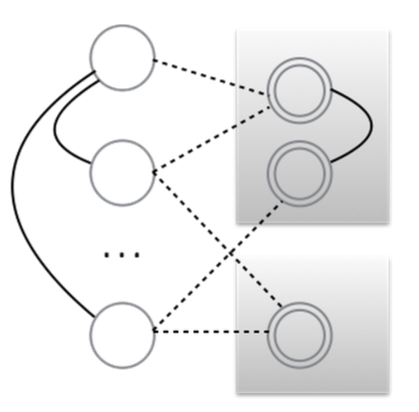} \\
      {\small Generic graph} &
      {\small Sparse graph} &
      {\small Dense graph} & 
      {\small Sparse+Dense graph}  \\
\end{tabular}}
\caption{Graph construction strategies.}
\label{fig:graph_model}
\end{figure}

The main ingredient for graph-based SSL approaches is the input graph itself. We demonstrate that the choice of graph construction mechanism has an important effect on the quality of SSL output. Depending on the edge link information as well as choice of vertex representation, there are multiple ways to create an input graph for SSL---(a) {\it Generic} graphs which represent observed neighborhood or link information connecting vertices (e.g., connections in a social network), (b) graphs constructed from {\it sparse feature representations} for each vertex (e.g., a bipartite Freebase graph connecting entity nodes with cell value nodes that capture properties of the entity occurring in a schema or table), (c) graphs constructed from {\it dense representations} for each vertex, i.e., use dense feature characteristics per node to define neighborhood (discussed in more detail in the next section), and (d) {\it augmented} graphs that use a mixture of the above.

Figure~\ref{fig:graph_model} shows an illustration of the various graph types. We focus on (b), (c) and (d) here since these are more applicable to natural language scenarios. Sparse instance-feature graphs (b) are typically provided as input for most SSL tasks in NLP. Next, we propose a method to automatically construct a graph (c) for text applications using semantic embeddings and use this to produce an augmented graph (d) that captures both sparse and dense per-vertex characteristics.

\subsection{Graph Augmentation with Dense Semantic Representations}
\label{sec:aug}
In the past, graph-based SSL methods have been widely applied to several NLP problems. In many scenarios, the nodes (and labels) represent textual information (e.g., query, document, entity name/type, etc.) and could be augmented with semantic information from the real world. Recently, some researchers have explored strategies to enhance the input graphs~\cite{KozarevaVT11} using external sources such as the Web or a knowledge base. However,
these methods require access to structured information from a knowledge base
or access to Web search results corresponding to a large number of targeted
queries from the particular domain. Unlike these methods, we propose a more robust strategy for graph augmentation that follows a two-step approach using only a large corpus of raw text. First,
we learn a dense vector representation that captures the underlying semantics
associated with each (text) node. We resort to recent state-of-the-art deep learning algorithms to
efficiently learn word and phrase semantic embeddings in a dense
low-dimensional space from a large text corpus using unsupervised methods.


We follow the recent work of Mikolov et al.~\cite{mikolov13a,mikolov13b} to compute continuous vector representations of words (or phrases) from very large datasets. The method takes a text corpus as input and learns a vector representation for every word (or phrase) in the vocabulary. We use the continuous skip-gram model~\cite{mikolov13a} combined with a hierarchical softmax layer in which each word in a sentence is used as an input to a log-linear classifier which tries to maximize classification of another word within the same sentence using the current word. More details about the deep learning architecture and training procedure can be found in~\cite{mikolov13a}. Moreover, these models can be efficiently parallelized and scale to huge datasets using a distributed training framework~\cite{DeanCMCDLMRSTYN12}. We obtain a 1000-dimensional vector representation (for each word) trained on 100 billion tokens of newswire text.\footnote{It is also possible to use pre-trained embedding vectors: https://code.google.com/p/word2vec/} For some settings (example dataset in Section~\ref{sec:exp}), nodes represent entity names (word collocations and not bag-of-words). We can train the embedding model to take this into account by treating entity mentions (e.g., within Wikipedia or news article text) as special words and applying the same procedure as earlier to produce embedding vectors for entities. Next, for each node $v = w_1 w_2 ... w_n$, we query the pre-trained vectors $\bf \mathcal{E}$ to obtain its corresponding embedding $v_{emb}$ from words in the node text.
\vspace{-.3cm}
\begin{align}
&v_{emb} =     \begin{cases}
     \mathcal{E}(v), & \text{if}\ v \in \mathcal{E} \\
      \frac{1}{n} \sum_i \mathcal{E}(w_i), & \text{otherwise}
    \end{cases}
\end{align}
\vspace{-.25in}


Following this, we compute a similarity function over pairs
of nodes using the embedding vectors, where $sim_{emb}(u, v) = u_{emb} \cdot
v_{emb}$. We filter out node pairs with low similarity values $<
\theta_{sim}$ and add an edge in the original graph $G = (V, E)$ for every
remaining pair.

Unfortunately, the above strategy requires $O(|V|^2)$ similarity computations which is infeasible in practice. To address this challenge, we resort to locality sensitive hashing (LSH)~\cite{Charikar02}, a random projection method used to efficiently approximate nearest neighbor lookups when data size and dimensionality is large. We use the node embedding vectors $v_{emb}$ and perform LSH to significantly reduce unnecessary pairwise computations that would yield low similarity values.\footnote{For LSH, we use $\theta_{sim}$=0.6, number of hash tables $D$=12, width $W$=10 in our experiments.}
\vspace{-.5em} 
\section{Experiments}
\label{sec:exp}
\setlength{\textfloatsep}{2pt}
\setlength{\dblfloatsep}{2pt}
\setlength{\dbltextfloatsep}{2pt}
\subsection{Experiment Setup}
\label{sec:expsetup}
\vspace{-0.1in}

\noindent {\bf Data:} We use two real-world datasets (publicly available from Freebase) for evaluation in this section.

\scalebox{0.72}{
\begin{tabular}{| c | c | c | c | c | c |}
\hline
\textbf{Data Name} & \textbf{Nodes} & \textbf{Edges} & \textbf{Labels} & \textbf{Avg, Deg.}\\
\hline
\hline
Freebase-Entity    & $301,638$ & $1,155,001$ & $192$ & $3.83$\\ \hline
Freebase-Relation  & $9,367,013$ & $16,817,110$ & $7,664$ & $1.80$\\ \hline
\end{tabular}}

Freebase-Entity (referred as FB-E) is the exact same dataset and setup used in previous works~\cite{TalukdarC14,Talukdar:ACL'10}. This dataset consists of cell value nodes and property nodes which are entities and Table properties in Freebase. An edge indicates that an entity appears in a table cell. The second dataset is Freebase-Relation (referred as
FB-R). This dataset comprises entity1-relation-entity2 triples from Freebase, which consists of more than 7000 relations and more than 8M triples. We extract two kinds of nodes from these triples, entity-pair nodes (e.g., {\it $<$Barack Obama, Hawaii$>$}) and entity nodes (e.g., {\it Barack Obama}). The former one is labeled with the relation type (e.g., {\it PlaceOfBirth}). An edge is created if two nodes have an entity in common.\\
\noindent {\bf Graph-based SSL systems:} We compare different graph-based SSL methods: ~{\bf \XPANDR}, both the vanilla method and the version that runs on the graph with semantic augmentation (as detailed in Section~\ref{sec:aug}), and ~{\bf \XPANDR-S}, the streaming approximation algorithm introduced in Section~\ref{sec:sparsity}. 

For baseline comparison, we consider two state-of-art existing works
\textbf{MAD}~\cite{Talukdar:ECML'09} and
\textbf{MAD-SKETCH}~\cite{TalukdarC14}. Talukdar and Pereira~\shortcite{Talukdar:ACL'10} show that MAD
outperforms traditional graph-based SSL algorithms.
MAD-SKETCH further approximates the label distribution on each node using Count-Min
Sketch to reduce the space and time complexity. To ensure a fair comparison, we obtained the MAD code directly from the authors and ran the exact same code on the same machine as EXPANDER for all experiments reported here. We obtained the same MRR performance (0.28) for MAD on the Freebase-Entity dataset (10 seeds/label) as reported by \cite{TalukdarC14}.\\
\noindent{\bf Parameters:} For the SSL objective function parameters, we set $\mu_1 = 1$, $\mu_2=0.01$ and $\mu_3=0.01$. We tried multiple settings for MAD and MAD-SKETCH algorithms and replicated the best reported performance metrics from \cite{TalukdarC14} using these values, so the baseline results are comparable to their system.\\
\noindent {\bf Evaluation:} Precision@K (referred as P@K) and Mean Reciprocal Rank (MRR) are used as evaluation metrics for all experiments, where higher is better. P@K measures the accuracy of the top
ranking labels (i.e., atleast one of the gold labels was found among the top K) returned by each method. MRR is calculated as $\frac{1}{|Q|} \sum_{v \in Q} \frac{1}{rank_v}$, where $Q \subseteq V$ is the test node set, and $rank_v$ is the rank of the gold label among the label distribution $\bf \hat{Y}_v$.

For experiments, we use the same procedure as reported in literature~\cite{TalukdarC14}, running each algorithm for 10 iterations per round (verified to be sufficient for convergence on these datasets) and then taking the average performance over 3 rounds. 
\vspace{-0.1in}
\subsection{Graph SSL Results}
\vspace{-0.1in}

\begin{table*}[ht!]
\centering
\scalebox{0.65}{
\begin{tabular}{| c | l | l | l | l | l || l | l | l | l | l |}
\hline
                 & \multicolumn{5}{c||}{5 seeds/label}  & \multicolumn{5}{c|}{10 seeds/label}    \\
\hline
\textbf{Methods} & \textbf{MRR} & \textbf{P@1} & \textbf{P@5} & \textbf{P@10} & \textbf{P@20} & \textbf{MRR} & \textbf{P@1} & \textbf{P@5} & \textbf{P@10} & \textbf{P@20}\\
\hline
\hline
MAD      & $0.2485$ & $0.1453$ & $0.3127$ & $0.4478$ & $0.5513$ & $0.2790$ & $0.1988$ & $0.3496$ & $0.4663$ & $0.5604$\\ \hline
\XPANDR & $0.3271$ & $0.2086$ & $0.4507$ & $0.6029$ & $0.7299$ & $0.3348$ & $0.1994$ & $0.4701$ & $0.6506$ & $0.7593$ \\ \hline
\XPANDR & $\textbf{0.3511}$ & $\textbf{0.2301}$ & $\textbf{0.4799}$ & $\textbf{0.6176}$ & $\textbf{0.7384}$ & $\textbf{0.3727}$ & $\textbf{0.23436}$ & $\textbf{0.5173}$ & $\textbf{0.6654}$ & $\textbf{0.7679}$ \\
(combined graph) & &  &  & &  & &  & & & \\ \hline
\end{tabular}}
\caption{Comparison of various graph transduction methods on the the Freebase-Entity graph.}
\label{table:5and10seeds}
\end{table*}
First, we quantitatively compare the graph-based SSL methods in terms
of MRR and Precision@K without considering the space and time complexity. Table~\ref{table:5and10seeds} shows the results with 5 seeds/label and 10 seeds/label on the Freebase-Entity dataset.

From the results, we have several findings: (1) Both \XPANDR-based methods
outperform MAD consistently in terms of MRR and Precison@K.
(2) Our algorithm on the enhanced graph using semantic embeddings (last row) produces significant gains over the original graph, which indicates that densifying the graph with additional information provides a useful technique for improving SSL in such scenarios.
\vspace{-0.1in}
\subsection{Streaming Sparsity versus Sketch}
\vspace{-0.1in}

\begin{table*}[ht!]
\centering
\scalebox{0.65}{
\begin{tabular}{| c | c | c | c | c | c | c | c |}
\hline
\textbf{Methods} & \textbf{MRR} & \textbf{P@1} & \textbf{P@5} & \textbf{P@10} & \textbf{P@20} & \textbf{compute time(s)} & \textbf{space(G)}\\
\hline
\hline
MAD      & $0.2485$ & $0.1453$ & $0.3127$ & $0.4478$ & $0.5513$ & $206.5$ & $9.10$\\ \hline
MAD-SKETCH (w=20 d=3)      & $0.2041$ & $0.1285$ & $0.2536$ & $0.3133$ & $0.4528$ & $30.0$ & $1.20$\\ \hline
MAD-SKETCH (w=109 d=3)     & $0.2516$ & $0.1609$ & $0.3206$ & $0.4266$ & $0.5478$ & $39.8$ & $2.57$\\ \hline\hline
\XPANDR & $0.3271$ & $0.2086$ & $0.4507$ & $0.6029$ & $0.7299$ & $256.4$ & $1.34$\\ \hline
\XPANDR-S (k=5) & $NA$ & $0.2071$ & $0.4209$ & $NA$ & $NA$ & $78.2$ & $0.62$\\ \hline
\XPANDR-S (k=10) & $NA$ & $0.2046$ & $0.4493$ & $0.5923$ & $NA$ & $94.0$ & $0.76$\\ \hline
\XPANDR-S (k=20) & $NA$ & $0.2055$ & $0.4646$ & $0.5981$ & $0.7221$ & $123.1.4$ & $0.82$\\ \hline
\end{tabular}}
\caption{Comparison of various scalable methods based on MAD and EXPANDER on the Freebase-Entity graph.}
\label{table:scalableEntity}
\end{table*}

\setlength{\dbltextfloatsep}{0.05in}
\setlength{\dblfloatsep}{0.05in}
\begin{table*}[ht!]
\centering
\scalebox{0.65}{
\begin{tabular}{| c | c | c | c | c | c | c | c |}
\hline
\textbf{Methods} & \textbf{MRR} & \textbf{P@1} & \textbf{P@5} & \textbf{P@10} & \textbf{P@20} & \textbf{compute time(s)} & \textbf{space(G)}\\
\hline \hline MAD-SKETCH (w=20 d=3)      & $0.1075$ & $0.0493$ & $0.21572$ &
$0.2252$ & $0.2902$ & $294$ & $12$\\ \hline\hline
\XPANDR-S (k=5) & $NA$ & $0.1054$ & $0.2798$ & $NA$ & $NA$ & $1092$ & $0.91$\\
\hline \XPANDR-S (k=10) & $NA$ & $0.1057$ & $0.2818$ & $0.3745$ & $NA$ &
$1302$ & $1.02$\\
\hline
\XPANDR-S (k=20) & $NA$ & $0.1058$ & $0.2832$ & $0.3765$ & $0.4774$ &
$1518$ & $1.14$\\
\hline
\end{tabular}}
\caption{Comparison of MAD and EXPANDER methods on the FB-R$_2$ graph, a subgraph of Freebase-Relation.} \label{table:scalableRelation}
\end{table*}

In this section, we compare the MAD-SKETCH and \XPANDR-S algorithms against the vanilla versions. The former one uses Count-Min Sketch to approximate the whole label distribution per node, while the latter uses streaming approximation to capture the sparsity of the label distribution. For Freebase-Entity dataset, we run these two methods with 5 seeds/label.\footnote{We observed similar findings with larger seed sizes.} The Freebase-Relation dataset is too big to run on a single machine, so we sample a smaller dataset FB-R$_2$\footnote{We create FB-R$_2$ by randomly picking 1000 labels and keeping only entity-pair nodes which belong to these labels and their corresponding edges (4.5M nodes, 7.4M edges).} from it. For this new dataset, we only compare the approximation methods MAD-SKETCH and our
\XPANDR-S, by picking 20 seeds/label and taking average over 3 rounds. We just
test MAD-SKETCH (w=20,d=3), since the setting MAD-SKETCH (w=109,d=3) runs out-of-memory using a single machine. We use protocol buffers (an efficient data serialization scheme) to
store the data for \XPANDR-S. For the space, we report the memory taken by the
whole process. For \XPANDR-S, as described in
section~\ref{sec:sparsity}, each node stores at most $k$ labels, so the MRR and
precision@K where $K > k$ are not available, and we refer it as NA.

Tables~\ref{table:scalableEntity},~\ref{table:scalableRelation} show results on Freebase-Entity and the smaller Freebase-Relation (FB-R$_2$) datasets, respectively. We make the following observations: (1)
MAD-SKETCH (w=109,d=3) can obtain similar performance compared with MAD, while
it can achieve about $5.2\times$ speedup, and $3.54\times$ space reduction. However, when
the sketch size is small (e.g. w=20,d=3), the algorithm loses quality in terms
of both MRR and Precision@K. For applications involving large label sizes, due to space limitations, we can only
allocate a limited memory size for each node, yet we should still be able to
retain the accurate or relevant labels within the available memory. On
FB-R$_2$ data, we observe that  MAD-SKETCH (w=109,d=3) is not executable, and
the MAD-SKETCH (w=20,d=3) yields poor results. (2) Comparing the \XPANDR~and
\XPANDR-S, the latter one obtains similar performance in terms of Precision@K,
while it achieves $3.28\times$ speedup for $k=5$ and $2.16\times$ space reduction.
Compared with the MAD-SKETCH, the speedup is not as steep mainly
because the algorithm needs to go through the tuple list and filter out the
ones below the threshold to ensure that we retain the ``good'' labels. However, we
can easily execute \XPANDR-S on the subset of Freebase-Relation
(FB-R$_2$), due to low space requirements ($\sim$$12\times$
lower than even MAD-SKETCH). Moreover, it outperforms MAD-SKETCH (w=20,d=3) in
terms of Precision@K.

{\bf Frequency Thresholding vs. Streaming Sparsity:}
We also compare our streaming approximation algorithm (\XPANDR-S) against a simple frequency-based thresholding technique (\FREQ) used often by online sparse learning algorithms (zero out small weights after each update Step 11 in Algorithm~\ref{algo:distributed}). However, this becomes computationally inefficient $O(degree * k)$ in our case especially for nodes with high degree, which is prohibitive since it requires us to aggregate label distributions from all neighbors before pruning.\footnote{We set threshold $\delta=0.001$ for \FREQ~based on experiments on a small heldout dataset.} In both cases, we can still maintain constant space complexity per-node by retaining only top-K labels after the update step. Table~\ref{table:sparsityFrequency} shows that the streaming sparsity approximation produces significantly better quality results (precision at 5, 10) than frequency thresholding in addition to being far more computationally efficient.

\begin{table}[ht!]
\centering
\setlength{\belowcaptionskip}{0.05in}
\scalebox{0.65}{
\begin{tabular}{| c | c | c | c |}
\hline
\textbf{Methods} & \textbf{P@1} & \textbf{P@5} & \textbf{P@10}\\
\hline \hline
\FREQ (k=5) & 0.1921 & 0.4066 & $NA$ \\
\XPANDR-S (k=5) & \bf 0.2071 & \bf 0.4209 & $NA$\\ \hline
\hline
\FREQ (k=10) & 0.2028 & 0.4216 & 0.5719 \\
\XPANDR-S (k=10) & \bf 0.2046 & \bf 0.4493 & \bf 0.5923 \\ \hline
\end{tabular}}
\caption{Comparison of sparsity approximation vs. frequency thresholding on Freebase-Entity dataset.} \label{table:sparsityFrequency}
\end{table}
\vspace{-0.1in}
\subsection{Graph SSL with Large Data, Label Sizes}
\vspace{-0.1in}

\begin{figure}[ht!]
\centering
\setlength{\dbltextfloatsep}{0.05in}
\setlength{\dblfloatsep}{0.05in}
\scalebox{0.72}{
\includegraphics[width=0.44\textwidth]{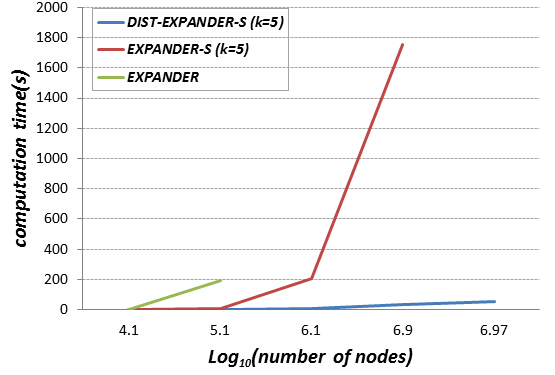}}
  \vspace{-1em}
\caption{Running time vs. Data size for single-machine versus distributed algorithm.}
  \vspace{-0.5em}
\label{fig:timevsszie}
\end{figure}

In this section, we evaluate the space and time efficiency of our distributed algorithm~\DXPANDR~(described in Section~\ref{sec:distributed}) coupled with the new streaming approximation update (Section~\ref{sec:sparsity}). To focus on large data settings, we only use Freebase-Relation data set in subsequent experiments. Following previous work~\cite{TalukdarC14}, we use the identity of each node as a label. In other words, the label size can potentially be as large as the size of the nodes $|V|$.
First, we test how the computation time scales with the number of nodes, by fixing label size. We follow a straightforward strategy: randomly sample different number of edges (and corresponding nodes) from the original graph. For each graph size, we randomly sample $1000$ nodes as seeds, and set their node identities as labels. We then run vanilla~\XPANDR, \XPANDR-S ($k=5$) and \DXPANDR-S ($k=5$) on each graph. The last one is a distributed version which partitions the graph and runs on $100$ machines. We show the running time for different node sizes in Figure~\ref{fig:timevsszie}.
\XPANDR~runs out-of-memory when the node size goes up to 1M, and the running time slows down significantly when graph size increases. \XPANDR-S (k=5) can handle all five data sets on a single machine and while the running time is better than \XPANDR, it starts to slow down noticeably on larger graphs with 7M nodes. \DXPANDR-S scales  quite well with the node size, and yields a 50-fold speedup when compared with \XPANDR-S when the node size is $\sim$7M.
\begin{figure}[ht!]
\centering
\scalebox{0.72}{
\includegraphics[width=0.4\textwidth]{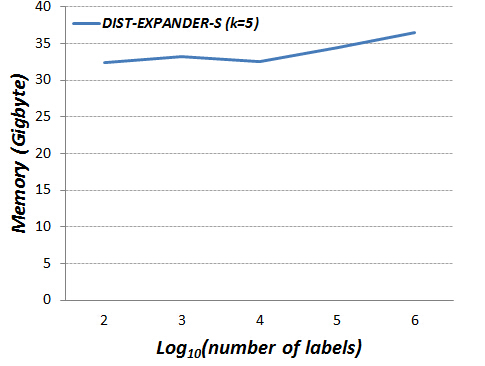}}
  \vspace{-1em}
\caption{Memory usage vs. Label size.}
  \vspace{-0.5em}
\label{fig:spacevssize}
\end{figure}
\begin{figure}[ht!]
\centering
\scalebox{0.8}{
\includegraphics[width=0.41\textwidth]{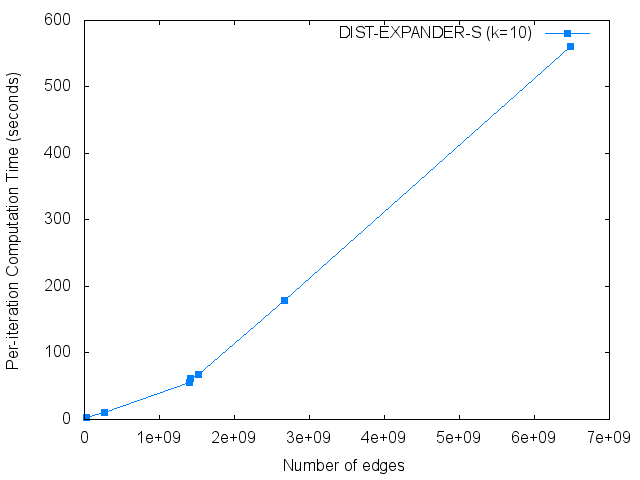}}
  \vspace{-1em}
\caption{Per iteration runtime on massive graphs of varying sizes distributed across 300 machines.}
\label{fig:distlarge}
\end{figure}

Figure~\ref{fig:spacevssize} illustrates how the memory usage scales with label size for our distributed version. For this scenario, we use the entire Freebase-Relation dataset and vary label size by randomly choosing different number of seed nodes as labels. We find that the overall space cost is consistently around 35GB, because our streaming algorithm captures a sparse constant space approximation of the label distribution and does not run out-of-memory even for large label sizes. Note that the distributed version consumes more than 30GB primarily because there will be redundant information recorded when partitioning the graph (including data replication for system fault-tolerance).

Finally, we  test how the distributed sparsity approximation algorithm scales on massive graphs with billions of nodes and edges. Since the Freebase graph is not sufficiently large for this setting, we construct graphs of varying sizes from a different dataset (segmented video sequences tagged with 1000 labels). Figure~\ref{fig:distlarge} illustrates that the streaming distributed algorithm scales very efficiently for such scenarios and runs quite fast. Each computing iteration for \DXPANDR-S runs to completion in just 2.3 seconds on a 17.8M node/26.7M edge graph and roughly 9 minutes on a much larger 2.6B node/6.5B edge graph.
\vspace{-.5em}

\section{Conclusion}
\label{sec:conc}
\vspace{-.5em}

Existing graph-based SSL algorithms usually require $O(m)$ space per node and do not scale to scenarios involving large label sizes $m$ and massive graphs. We propose a novel streaming algorithm that effectively and accurately captures the sparsity of the label distribution. The algorithm operates efficiently in a streaming fashion and reduces the space
complexity per node to $O(1)$ in addition to yielding high quality performance. Moreover, we extend the method with a distributed algorithm that scales elegantly to large data and label sizes (for example,
billions of nodes/edges and millions of labels). We also show that graph augmentation using unsupervised learning techniques can provide a robust strategy to yield performance gains for SSL problems involving natural language.
\vspace{-.5em}

\subsubsection*{Acknowledgements}
We thank Partha Talukdar for useful pointers to MAD code and Kevin Murphy for providing us access to the Freebase-Relation dataset.

\bibliographystyle{abbrv}
\balance
\bibliography{expander}

\begin{thebibliography}{10}

\bibitem{ragrawal2013}
R.~Agrawal, A.~Gupta, Y.~Prabhu, and M.~Varma.
\newblock Multi-label learning with millions of labels: Recommending advertiser
  bid phrases for web pages.
\newblock In {\em Proceedings of the International World Wide Web Conference},
  2013.

\bibitem{Giraph}
Apache giraph.
\newblock \url {http://giraph.apache.org/}, 2013.

\bibitem{Baluja:WWW'08}
S.~Baluja, R.~Seth, D.~Sivakumar, Y.~Jing, J.~Yagnik, S.~Kumar,
  D.~Ravichandran, and M.~Aly.
\newblock Video suggestion and discovery for {Y}outube: Taking random walks
  through the view graph.
\newblock In {\em Proceedings of the 17th International Conference on World
  Wide Web}, WWW '08, pages 895--904, 2008.

\bibitem{Belkin05}
M.~Belkin, P.~Niyogi, and V.~Sindhwani.
\newblock On manifold regularization.
\newblock In {\em Proceeding of the Conference on Artificial Intelligence and
  Statistics (AISTATS)}, 2005.

\bibitem{Bengio+al-ssl-2006}
Y.~Bengio, O.~Delalleau, and N.~{Le Roux}.
\newblock Label propagation and quadratic criterion.
\newblock In O.~Chapelle, B.~Sch{\"o}lkopf, and A.~Zien, editors, {\em
  Semi-Supervised Learning}, pages 193--216. {MIT} Press, 2006.

\bibitem{blitzer-zhu:2008:ACLTutorials}
J.~Blitzer and X.~J. Zhu.
\newblock Semi-supervised learning for natural language processing.
\newblock In {\em ACL-HLT Tutorial}, June 2008.

\bibitem{CarlsonBKSHM10}
A.~Carlson, J.~Betteridge, B.~Kisiel, B.~Settles, E.~R.~H. Jr., and T.~M.
  Mitchell.
\newblock Toward an architecture for never-ending language learning.
\newblock In {\em AAAI}, 2010.

\bibitem{ChaSchZie06}
O.~Chapelle, B.~Sch{\"o}lkopf, and A.~Zien, editors.
\newblock {\em Semi-Supervised Learning}.
\newblock MIT Press, Cambridge, MA, 2006.

\bibitem{Charikar02}
M.~Charikar.
\newblock Similarity estimation techniques from rounding algorithms.
\newblock In {\em Proceedings of the thiry-fourth annual ACM symposium on
  Theory of computing}, pages 380--388, 2002.

\bibitem{Cormode:2005}
G.~Cormode and S.~Muthukrishnan.
\newblock An improved data stream summary: The count-min sketch and its
  applications.
\newblock {\em Journal of Algorithms}, 55(1):58--75, 2005.

\bibitem{DasSmith:2012}
D.~Das and N.~A. Smith.
\newblock Graph-based lexicon expansion with sparsity-inducing penalties.
\newblock In {\em Proceedings of the 2012 Conference of the North American
  Chapter of the Association for Computational Linguistics: Human Language
  Technologies}, pages 677--687. Association for Computational Linguistics,
  2012.

\bibitem{DeanCMCDLMRSTYN12}
J.~Dean, G.~Corrado, R.~Monga, K.~Chen, M.~Devin, Q.~V. Le, M.~Z. Mao,
  M.~Ranzato, A.~W. Senior, P.~A. Tucker, K.~Yang, and A.~Y. Ng.
\newblock Large scale distributed deep networks.
\newblock In {\em Proceedings of NIPS}, pages 1232--1240, 2012.

\bibitem{imagenet_cvpr09}
J.~Deng, W.~Dong, R.~Socher, L.-J. Li, K.~Li, and L.~Fei-Fei.
\newblock {ImageNet: A Large-Scale Hierarchical Image Database}.
\newblock In {\em CVPR09}, 2009.

\bibitem{VandurmeLall09}
B.~V. Durme and A.~Lall.
\newblock Streaming pointwise mutual information.
\newblock In Y.~Bengio, D.~Schuurmans, J.~Lafferty, C.~Williams, and
  A.~Culotta, editors, {\em Advances in Neural Information Processing Systems
  22}, pages 1892--1900. 2009.

\bibitem{Goyal:NAACL'09}
A.~Goyal, H.~Daum{\'e}, III, and S.~Venkatasubramanian.
\newblock Streaming for large scale {NLP}: Language modeling.
\newblock In {\em Proceedings of Human Language Technologies: The 2009 Annual
  Conference of the North American Chapter of the Association for Computational
  Linguistics}, NAACL '09, pages 512--520, 2009.

\bibitem{Joachims:1999}
T.~Joachims.
\newblock Transductive inference for text classification using support vector
  machines.
\newblock In {\em Proceedings of the Sixteenth International Conference on
  Machine Learning}, pages 200--209, 1999.

\bibitem{Joachims03}
T.~Joachims.
\newblock Transductive learning via spectral graph partitioning.
\newblock In {\em Proceedings of ICML}, pages 290--297, 2003.

\bibitem{kowalski09}
M.~Kowalski and B.~Torr{\'e}sani.
\newblock {Sparsity and persistence: mixed norms provide simple signal models
  with dependent coefficients}.
\newblock {\em {Signal, Image and Video Processing}}, 3(3):251--264, Sept.
  2009.

\bibitem{KozarevaVT11}
Z.~Kozareva, K.~Voevodski, and S.-H. Teng.
\newblock Class label enhancement via related instances.
\newblock In {\em Proceedings of EMNLP}, pages 118--128, 2011.

\bibitem{Pregel}
G.~Malewicz, M.~H. Austern, A.~J. Bik, J.~C. Dehnert, I.~Horn, N.~Leiser, and
  G.~Czajkowski.
\newblock Pregel: a system for large-scale graph processing.
\newblock In {\em Proceedings of the 2010 ACM SIGMOD International Conference
  on Management of data}, pages 135--146, 2010.

\bibitem{Manku:VLDB'02}
G.~S. Manku and R.~Motwani.
\newblock Approximate frequency counts over data streams.
\newblock In {\em Proceedings of the 28th International Conference on Very
  Large Data Bases}, VLDB '02, pages 346--357, 2002.

\bibitem{mikolov13a}
T.~Mikolov, K.~Chen, G.~Corrado, and J.~Dean.
\newblock Efficient estimation of word representations in vector space.
\newblock In {\em Proceedings of Workshop at ICLR}, 2013.

\bibitem{mikolov13b}
T.~Mikolov, I.~Sutskever, K.~Chen, G.~Corrado, and J.~Dean.
\newblock Distributed representations of words and phrases and their
  compositionality.
\newblock In {\em Proceedings of NIPS}, 2013.

\bibitem{Osborne:ACL'14}
M.~Osborne, A.~Lall, and B.~V. Durme.
\newblock Exponential reservoir sampling for streaming language models.
\newblock In {\em Proceedings of The 52nd Annual Meeting of the Association for
  Computational Linguistics}, ACL '2014, pages 687--692, 2014.

\bibitem{Seeger01}
M.~Seeger.
\newblock Learning with labeled and unlabeled data.
\newblock Technical report, 2001.

\bibitem{SubramanyaB09}
A.~Subramanya and J.~A. Bilmes.
\newblock Entropic graph regularization in non-parametric semi-supervised
  classification.
\newblock In {\em Proceedings of NIPS}, pages 1803--1811, 2009.

\bibitem{Subramanya:EMNLP'10}
A.~Subramanya, S.~Petrov, and F.~Pereira.
\newblock Efficient graph-based semi-supervised learning of structured tagging
  models.
\newblock In {\em Proceedings of the 2010 Conference on Empirical Methods in
  Natural Language Processing}, EMNLP '10, pages 167--176, 2010.

\bibitem{TalukdarC14}
P.~Talukdar and W.~Cohen.
\newblock Scaling graph-based semi supervised learning to large number of
  labels using count-min sketch.
\newblock In {\em Proceedings of AISTATS}, pages 940--947, 2014.

\bibitem{Talukdar:ECML'09}
P.~P. Talukdar and K.~Crammer.
\newblock New regularized algorithms for transductive learning.
\newblock In {\em Proceedings of the European Conference on Machine Learning
  and Knowledge Discovery in Databases: Part II}, ECML PKDD '09, pages
  442--457, 2009.

\bibitem{Talukdar:ACL'10}
P.~P. Talukdar and F.~Pereira.
\newblock Experiments in graph-based semi-supervised learning methods for
  class-instance acquisition.
\newblock In {\em Proceedings of the 48th Annual Meeting of the Association for
  Computational Linguistics}, ACL '10, pages 1473--1481, 2010.

\bibitem{TalukdarRPRBP08}
P.~P. Talukdar, J.~Reisinger, M.~Pasca, D.~Ravichandran, R.~Bhagat, and
  F.~Pereira.
\newblock Weakly-supervised acquisition of labeled class instances using graph
  random walks.
\newblock In {\em EMNLP}, pages 582--590, 2008.

\bibitem{tibshirani96}
R.~Tibshirani.
\newblock Regression shrinkage and selection via the lasso.
\newblock {\em Journal of the Royal Statistical Society (Series B)},
  58:267--288, 1996.

\bibitem{Tommi03}
A.~C. Tommi and T.~Jaakkola.
\newblock On information regularization.
\newblock In {\em Proceedings of the 19th UAI}, 2003.

\bibitem{Ugander:2013}
J.~Ugander and L.~Backstrom.
\newblock Balanced label propagation for partitioning massive graphs.
\newblock In {\em Proceedings of the Sixth ACM International Conference on Web
  Search and Data Mining}, pages 507--516, 2013.

\bibitem{VanDurme:ACL'11}
B.~Van~Durme and A.~Lall.
\newblock Efficient online locality sensitive hashing via reservoir counting.
\newblock In {\em Proceedings of the 49th Annual Meeting of the Association for
  Computational Linguistics: Human Language Technologies: Short Papers - Volume
  2}, HLT '11, pages 18--23, 2011.

\bibitem{WangJC13}
Y.~Wang, R.~Ji, and S.-F. Chang.
\newblock Label propagation from imagenet to 3d point clouds.
\newblock In {\em Proceedings of CVPR}, pages 3135--3142. IEEE, 2013.

\bibitem{Zhu05}
X.~Zhu.
\newblock Semi-supervised learning literature survey.
\newblock Technical Report 1530, Computer Sciences, University of
  Wisconsin-Madison, 2005.

\bibitem{Zhu03ssl}
X.~Zhu, Z.~Ghahramani, and J.~Lafferty.
\newblock Semi-supervised learning using gaussian fields and harmonic
  functions.
\newblock In {\em Proceedings of ICML}, pages 912--919, 2003.

\end{thebibliography}

\end{document}